\documentclass{llncs}

\usepackage{amssymb}
\usepackage{hyperref}
\usepackage{multicol}
\usepackage{multirow}
\usepackage{graphicx}
\usepackage{pgfplots}
\pgfplotsset{compat=1.18} 
\usepackage{fancyhdr}
\usepackage{xcolor}
\usepackage{tcolorbox}
\usepackage{tabularx,booktabs}

\usepackage{dingbat}
\usepackage{float}
\usepackage{listings}
\usepackage{amsmath}
\usepackage{xspace}

\lstdefinestyle{mypython}{
  language=Python,
  basicstyle=\ttfamily\small,
  keywordstyle=\color{blue}\bfseries,
  commentstyle=\color{gray},
  stringstyle=\color{magenta},
  showstringspaces=false,
  breaklines=true,
  frame=single,
  numbers=left,
  numberstyle=\tiny\color{gray},
  captionpos=b
}

\newcommand{\NBZ}{\mathcal{Z}\xspace}
\newcommand{\NbNoeudsCentraux}{\mathcal{N}\xspace}

\begin{document}
\renewcommand{\labelenumii}{\arabic{enumi}.\arabic{enumii}}

\title{GPML: Graph Processing for Machine Learning}

\author{Majed Jaber\inst{1,2} \and
Julien Michel\inst{1,2} \and
Nicolas Boutry\inst{2} \and
Pierre Parrend\inst{1,2}}
\authorrunning{M. Jaber et al.}
%
\institute{EPITA, Laboratoire de Recherche de l’EPITA (LRE), 14-16 Rue Voltaire, 94270 Le Kremlin Bicêtre
\email{\{julien.michel, nicolas.boutry, pierre.parrend\}@epita.fr}
\and
ICube UMR7357, Université de Strasbourg, CNRS, F-67000 Strasbourg, France
\email{majed.jaber@etu.unistra.fr}}

\maketitle

\begin{abstract}
The dramatic increase of complex, multi-step, and rapidly evolving attacks in dynamic networks involves advanced cyber-threat detectors. The GPML (Graph Processing for Machine Learning) library addresses this need by transforming raw network traffic traces into graph representations, enabling advanced insights into network behaviors. The library provides tools to detect anomalies in interaction and community shifts in dynamic networks. GPML supports community and spectral  metrics extraction, enhancing both real-time detection and historical forensics analysis. This library supports modern cybersecurity challenges with a robust, graph-based approach.
\end{abstract}


\section{Metadata}
\label{sec:metadata}

\begin{table}[H]
\scriptsize
\begin{tabular}{|l|p{6.5cm}|p{6.5cm}|}
\hline
\textbf{Nr.} & \textbf{Code metadata description} & \textbf{Please fill in this column} \\
\hline
C1 & Current code version & \v1.0.0 \\
\hline
C2 & Permanent link to code/repository used for this code version &
 \url{https://github.com/lre-security-systems-team/gpml} \\
\hline
C3  & Permanent link to Reproducible Capsule & \url{https://www.kaggle.com/code/majedjaber/gpml-reproducible-capsule}\\
\hline
C4 & Legal Code License   & ISC  \url{https://opensource.org/license/isc-license-txt} \\
\hline
C5 & Code versioning system used & git \\
\hline
C6 & Software code languages, tools, and services used & python \\
\hline
C7 & Compilation requirements, operating environments \& dependencies & 
\url{https://github.com/lre-security-systems-team/gpml/blob/main/requirements.txt}
 \\
\hline
C8 & If available Link to developer documentation/manual & \url{https://github.com/lre-security-systems-team/gpml/blob/main/README.md} \\
\hline
C9 & Support email for questions & 
\{julien.michel,majed.jaber\}@epita.fr\\
\hline
\end{tabular}
\caption{METADATA}
\label{codeMetadata} 
\end{table}

\section{Motivation and significance}

In today's digital landscape, network security faces growing challenges due to the increasing complexity of cyber-threats ~\cite{uma2013survey}. Attackers constantly improve their techniques, targeting vulnerabilities across interconnected devices, systems, and services. Traditional security measures often struggle to keep up, as they primarily rely on signature-based detection~\cite{kothamali2022limitations} or rule-based methods~\cite{sarker2024multi}, which may not capture emerging threats in network traffic. To effectively monitor, analyze, and secure network environments, new approaches are required—approaches that can handle large-scale and dynamic data from highly interconnected environments while addressing the alert fatigue in security operations \cite{tariq2025alert}.

Two main approaches emerge from the literature: graph analytics, or complex network analysis, which characterizes the connectivity properties on the graph itself, and embedding analysis, which extracts information about node surroundings for each node. Graph analytics leverages Community, Spectral, or Complex Network information to quantify the relative connectivity of node groups, the structure of the connections between nodes, or the position wrt. to whole network. Embedding analysis typically rely on Graph Neural Networks (GNN) and their variants. 
In all cases, graph-based models represent entities as nodes and their interactions as edges and enable effective analysis of network traffic.
Graph Neural Networks (GNNs) extend this by learning at the node, edge, and graph levels~\cite{wu2020comprehensive}. While traditional GNNs rely on node features and topology~\cite{liu2020graphsage}, recent models integrate edge features to better capture interaction-specific information such as packet volume or connection type. Edge-aware variants like E-GraphSage~\cite{lo2022graphsage} and NE-GConv ~\cite{altaf2023ne} incorporate these features during aggregation, improving the precision of edge-level predictions. NE-GConv performs binary edge classification, whereas E-GraphSage supports both binary and multi-class outputs.

Dynamic Graph Community (DGC) metrics significantly enhance detection performance in network traffic classification tasks. Enriching the baseline feature set, it captures dynamic interaction patterns in network data.
The spectral method in GPML library ($SPEC- TRA$) analyzes graph states over time using time windows and classifies attacks based on aggregated behaviors. While GNNs aim to identify specific malicious edges, $SPECTRA$ detects broader anomalous evolutions over multiple attack categories. 

GPML library integrates both community and spectral analytics to leverage a graph-based approach to network security analysis. This library transforms raw network traffic data into graph representations~\cite{jaber2024graph}. This approach not only enhances the detection of known issues but also reveals hidden or emerging patterns within complex networks. By identifying interconnections through communities and spectral analysis, and monitoring their changes over time, it becomes possible to detect unusual connectivity patterns, isolate suspicious nodes, and proactively respond to potential threats. Tracking both static and dynamic network metrics enables both real-time and historical analyses are critical.
GPML library builds on established methods in network graph analysis and leverages widely used libraries like NetworkX ~\cite{hagberg2020networkx} for graph operations and Pandas ~\cite{mckinney2011pandas} for data handling.


In future studies, the GPML library could serve as a foundational tool for developing AI-driven solutions to detect sophisticated threats. 

\section{Software description}



\subsection{Supported metrics}


The graph community metrics are calculated from a graph community partition at time $t$, and their dynamicity is calculated from the difference between time $t+1$ and $t$. Let $V_t$ be the number of node of a community at time $t$. \textbf{Stability}\(=\frac{|V_t\cap V_{t+1}|- |(V_t\cap\bar{V}_{t+1})\cup(V_{t+1}\cap\bar{V}_t)|}{|V_t\cup V_{t+1}|} \), is a ratio of similarity between two consecutive states of a community. \textbf{Density} in a community is the probability for a node to be adjacent to any given node in the community, \textbf{Conductance} is the proportion of communications pointing outside the community and \textbf{Degree} are the number of edges going out of a node. Refer to \cite{10.1145/2350190.2350193} for their definition.

The spectral metrics are derived from the spectrum $\Lambda\_t$ of the Laplacian matrix at time $t$, for more details you can refer to our work about detecting attacks using spectral graph analysis \cite{jaber2024graph}. We denote by $\Lambda\_t[i]$ the $i^{th}$ eigenvalue, $i \in [1,n]$, sorted in increasing order, and by $\NBZ(t)$ the multiplicity of zero in $\Lambda\_t$. \textbf{Connectedness}=($\exp{\left(1/\NBZ(t) - 1\right)}$), measures interconnectivity in the network. Let ($\NbNoeudsCentraux$) be the number of network devices (e.g. switches and servers). \textbf{Flooding}=$((\frac{1}{\NbNoeudsCentraux} \sum_{i=\NBZ(t)+1}^{\NBZ(t) + \NbNoeudsCentraux} \Lambda\_t[i]) - 1)$ and \textbf{Wiriness}=$(\frac{1}{\NbNoeudsCentraux} \sum_{i=n - \NbNoeudsCentraux + 1}^{n} \Lambda\_t[i])$, 

\subsection{Software architecture}

\begin{figure}[h]
    \centering
    \begin{minipage}{0.7\textwidth}
        \centering
        \includegraphics[width=\textwidth]{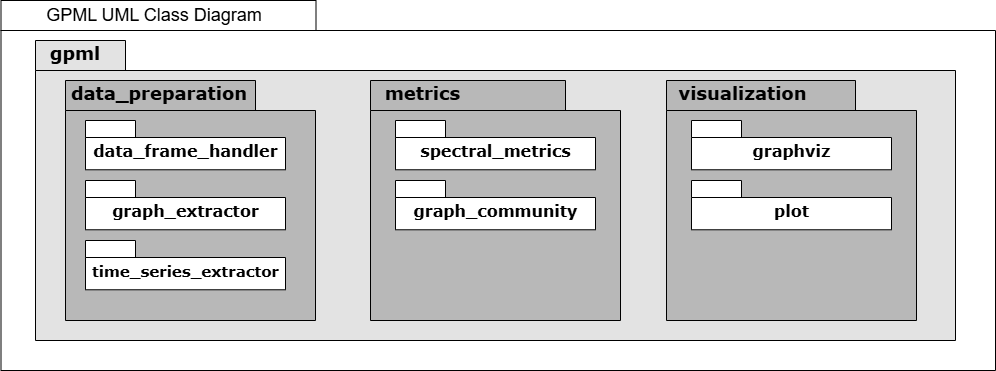}
        \caption{UML diagram showing the structure of the GPML library }
        \label{fig.gpml_components}
    \end{minipage}
\end{figure}

The library is organized into three main components, as illustrated in Fig~\ref{fig.gpml_components}. The hierarchy starts with the root \textit{gpml} directory, branching down into three primary sub-directories: \textit{data\_preparation}, \textit{metrics}, and \textit{visualization}.
\begin{itemize}
    \item The \textit{data\_preparation} directory is responsible for preparing data, starting with CSV files. It contains three classes: \textit{data\_frame\_handler.py}, \textit{graph\_extractor.py} and \textit{time\_series\_extractor.py}

    \item The second directory, \textit{metrics}, is central to our methodology, and computes spectral metrics in \textit{spectral\_metrics.py} as well as community metrics in \textit{graph\_community.py}.
    
    \item The final directory, \textit{visualization}, focuses on presenting graphs of nodes and edges for user interpretation, using network traffic CSV files as input. Visualization can be done through \textit{plot.py}, which extracts and plots graphs using the \textit{networkx} library, or \textit{graphviz.py}, which enhances representation and provides interactivity for HTML web versions.
\end{itemize}

\subsection{Software Implementation}
The library 
is structured under its main directory as follows: \begin{itemize}
\item gpml - Contains the core processing modules for two primary methodologies: community and spectral graph detection, along with components for graph processing and visualization. This is the central part of the library, offering diverse functionalities that aid in attack classification and dataset exploration.
\item data - Includes various datasets in CSV format.
\item doc - Holds the library documentation, detailing the dataset functionalities and usage.
\item test - Contains test cases for regression, providing examples that can be adapted to similar datasets using the specified constraints and parameters. 
\end{itemize}
\begin{figure}[H]
    \centering
    \begin{minipage}{0.7\textwidth}
        \centering
        \includegraphics[width=\textwidth]{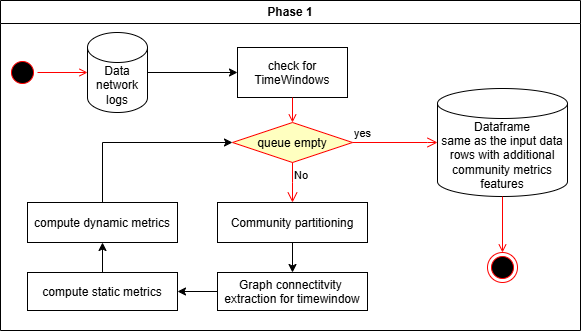}
        \caption{UML workflow diagram community strategy}
        \label{fig.uml_workflow_community_graph}
    \end{minipage}
\end{figure}

\subsection{Software functionalities}
The library functionalities can be divided into three main parts:
\begin{itemize}
    \item Extracting community graph features,
    \item Extracting spectral graph features,
    \item Plotting connectivity graphs.
\end{itemize}

Additionally, because the library requires correct inputs to function normally, you must make sure your dataset contains the following common features that exist in every traffic network data:
\begin{itemize}
    \item Timestamp for the arriving packets.
    \item Source and destination IP addresses.
    \item For spectral metrics extraction, you need in addition to the above, the
total number of packets, size of bytes and the rate of packets. These features exists in every traffic network data logs.
\end{itemize}

\begin{figure}[H]
    \centering
    \begin{minipage}{0.7\textwidth}
        \centering
        \includegraphics[width=\textwidth]{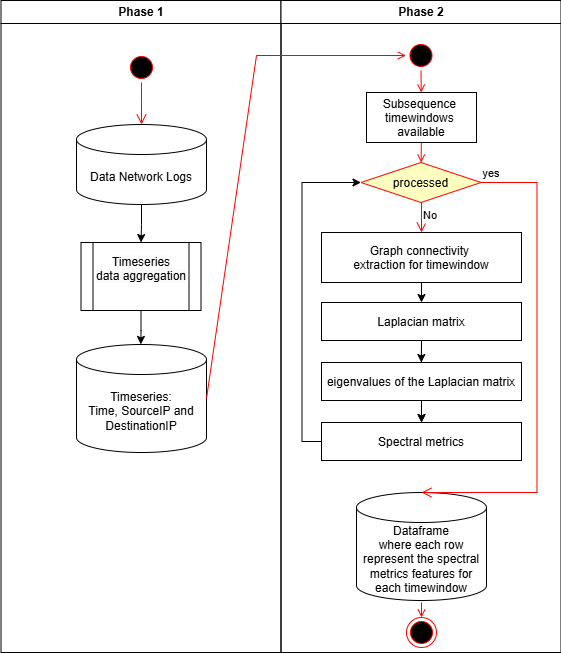}
        \caption{UML workflow diagram for spectral graph strategy}
        \label{fig.uml_workflow_spectral_graph}
    \end{minipage}
\end{figure}

The library provides a set of functionalities that help to add new features for better predictions. The main functionalities are:

\begin{lstlisting}[style=mypython, caption={Extraction of time series}]
time_series_extractor(df,stime,time_unit,features_list, sortby_list,groupby_list,aggregation_dict)
\end{lstlisting}
     
    In the \textit{time\_series\_extractor}, a time interval of \textit{$t=1s$} is applied to the dataframe \textit{df} provided by the user. Within this interval, the features are sorted according to \textit{sortby\_list} and grouped based on the features specified in the \textit{groupby\_list}. The remaining features, such as packets, bytes and rates are aggregated using the functions defined in the \textit{aggregation\_dict} such as mean, avg and max/min functions. This transformation converts the dataset into a time series format, typically reducing the number of rows compared to the original dataset.

\begin{lstlisting}[style=mypython, caption={Insertion of graph community metrics}]
insert_graph_community_metrics(dataframe,time_interval, date_time,edge_source,edge_dest,label,name,date_timestamp, community_strategy,continuity)
\end{lstlisting}
    The \textit{insert\_metrics\_to\_dataframe} function take a \textit{pandas DataFrame} and for a given time windows and community partitioning strategy as the parameters \textit{time\_interval} and \textit{community\_strategy} will produce the corresponding community metrics and insert them back to the \textit{DataFrame} as columns. The user needs to specify the column used for the time as \textit{date\_time} and the columns used for the nodes as lists in the \textit{edge\_source} and \textit{edge\_dest} parameters. A column from the \textit{DataFrame} has to be set to \textit{label} for the edges and a \textit{name} for a suffix of the new columns in the returned \textit{DataFrame}. A \textit{continuity} parameter has been added to choose if the user allows holes in the timeline of the data. The input \textit{DataFrame} is divided in time windows, for each time windows primarily, three underlying method will be called by this function: \textit{gc\_metrics\_first\_order(G)} which calculates metrics by one travel of the graph \textit{G}, \textit{gc\_metrics\_second\_order(forder\_metrics\_c, forder\_metrics\_g)} which calculates metrics by one travel over the first order metrics, and \textit{propagate\_communities(g1, g2, center, center\_t)} which creates the correspondence of communities from two graph at consecutive time windows. 
    Dynamic community metrics are then computed from those metrics and propagated through communities; all of them are added to the output \textit{DataFrame}.

\begin{lstlisting}[style=mypython, caption={Extraction of spectral metrics}]
spectral_metrics_extractor(ts,stime,saddr,daddr,pkts,
bytes_size,rate,lbl_category)
\end{lstlisting}
    In the \textit{spectral\_metrics\_extractor}, the parameters are provided by the user; however, constructing weighted edges within the network graph extracted over each time window requires selecting a specific feature. For this purpose, \textit{pkts}, \textit{bytes\_size}, and \textit{rate} are passed as inputs to weigh the graph during processing across three distinct topologies within each time window. Spectral metrics are then computed for each time window at the midpoint and the end. The output of this function is a new dataframe where each row represents a time window, including its corresponding common features, spectral features, and the associated label.

\begin{lstlisting}[style=mypython, caption={Print graph}]
print_graph(dataset,graph_type,label,src_addr,dst_addr, sport,dport,url,title,attack_name,src_mac,dst_mac)
\end{lstlisting}
    The \textit{print\_graph} function is designed to display the graph in two formats: a non-interactive format using \textit{networkx} and an interactive \textit{HTML} format that allows user interactions.

\section{Illustrative examples}

\subsection{Code snippets}    
\begin{figure}[h]
    \centering
    \begin{minipage}{0.7\textwidth}
        \centering
        \includegraphics[width=\textwidth]{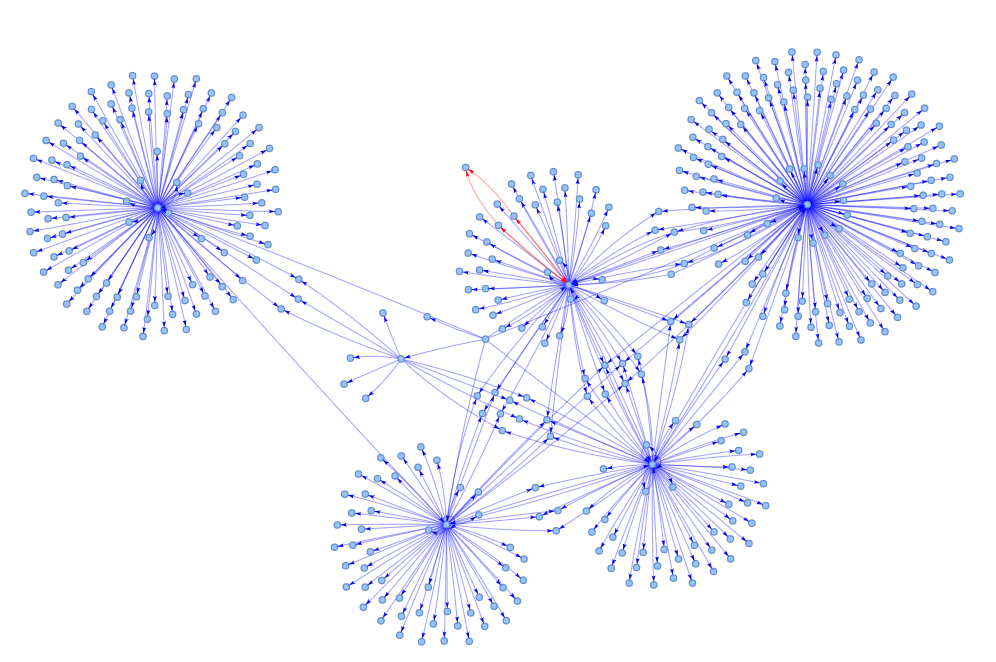}
        \caption{Ransomware attack in Ton-IoT dataset presented via HTML using graphviz function that exist in GPML library}
        \label{fig.ton_rensomware_graph_presetation}
    \end{minipage}
\end{figure}

\begin{itemize}
    \item Community metrics example


\begin{lstlisting}[style=mypython, caption={Insertion of graph community metrics - Example}]
from datetime import date, timedelta, datetime
from gpml.data_preparation import data_frame_handler as ins_data
community_df = ins_data.insert_graph_community_metrics(df, timedelta(minutes=5),'Date time',['Source IP'], ['Destination IP'],'Label', 'ip5',date_timestamp=False, community_strategy='louvain',continuity=True)

\end{lstlisting}    
    \item Spectral metrics example

    - Parameters:
    
\begin{lstlisting}[style=mypython, caption={Spectral metrics - parameters}]
features = ['stime', 'datetime', 'saddr', 'daddr', 'sport', 'dport', 'pkts', 'bytes', 'rate', 'attack', 'category', 'subcategory', 'weight', 'dur', 'mean', 'sum', 'min', 'max','spkts', 'dpkts', 'srate', 'drate']
sortby_list = ['stime']
groupby_list = ['stime', 'datetime', 'saddr', 'daddr']
aggregation_dict = { 'pkts': 'sum', 'bytes': 'sum', 'attack': 'first', 'category': 'first', 'subcategory': 'first', 'rate': 'mean', 'dur': 'mean', 'mean': 'mean', 'sum': 'mean', 'min': 'mean', 'max': 'mean', 'spkts': 'mean', 'srate': 'mean', 'drate': 'mean', 'weight': 'sum'
}
\end{lstlisting}  

-Timeseries extraction:

\begin{lstlisting}[style=mypython, caption={Extraction of time series - Example}]
ts = time_series_extractor(df, 'stime', 's', features_list, sortby_list, groupby_list, aggregation_dict)
\end{lstlisting}
    
-Extracting spectral metrics:
\begin{lstlisting}[style=mypython, caption={Extraction of spectral metrics - Example}]
spectral_metrics_df = spectral_metrics_extractor(ts, stime, saddr, daddr, pkts, bytes, rate, lbl_category)
print(res)
\end{lstlisting}

    \item Graph visualization example

\begin{lstlisting}[style=mypython, caption={Upload data - Example}]
import pandas as pd
df = pd.read_csv('data/ton_iot/ransomware-normal.csv')
print_graph(df, 'ip', 'label', 'src_ip', 'dst_ip', src_port', 'dst_port')
            
\end{lstlisting}
    The output file is saved in $/graph\_representation/$ directory as an $.html$ extention file as shown in Fig.~\ref{fig.ton_rensomware_graph_presetation}
\end{itemize}

\subsection{Evaluation and effectiveness}   

\begin{figure}[H]
\caption{Comparison of graph community approaches with baseline on UGR16 dataset with 5-folds evaluation using XGboost. Base set is original dataset feature space, the other one are the same dataset enriched with incrementally: graph metrics, graph community metrics and dynamic graphe community metrics. }
\label{fig:community_evaluation}
\begin{tabular}{ccc}
\multicolumn{3}{c}{} \\
\resizebox{0.33\linewidth}{!}{\begin{tikzpicture}
\begin{axis}[xlabel={Binary prediction},
ylabel={Score},
xmin=0,
xmax=3,
ymin=0.4,
ymax=1,
xtick={0,1,2,3},
xticklabels={Base set, Graph, Graph community, DGC},
ytick={0.5, 0.6, 0.7, 0.8, 0.9, 1},
legend pos=south east,
legend style={fill=none},
ymajorgrids=true,
grid style=dashed]
\addplot[
,color=blue,
mark=square,
]
coordinates {
(0,0.5218753042087866)(1,0.5652289759073101)(2,0.9519196161088365)(3,0.9564238801325787)
};
\addlegendentry{MCC}
\addplot[
,color=orange,
mark=square,
]
coordinates {
(0,0.7198059877883825)(1,0.7375957967065678)(2,0.9601868969428224)(3,0.9639476383291294)
};
\addlegendentry{Balanced Accuracy}
\addplot[
,color=green,
mark=square,
]
coordinates {
(0,0.44252751329742307)(1,0.47768817860373697)(2,0.9205236250285058)(3,0.9280337846719465)
};
\addlegendentry{TPR}
\addplot[
,color=cyan,
mark=square,
]
coordinates {
(0,0.9970844622793418)(1,0.997503414809399)(2,0.9998501688571393)(3,0.9998614919863122)
};
\addlegendentry{TNR}
\end{axis}
\end{tikzpicture}}
\resizebox{0.33\linewidth}{!}{\begin{tikzpicture}
\begin{axis}[xlabel={multi-class Prediction},
ylabel={F1\_Score},
xmin=0,
xmax=3,
ymin=0,
ymax=1,
xtick={0,1,2,3},
xticklabels={Base set, Graph, Graph community, DGC},
ytick={0.2, 0.4, 0.6, 0.8, 1},
legend pos=south east,
legend style={fill=none},
ymajorgrids=true,
grid style=dashed]
\addplot[
,color=green,
mark=square,
]
coordinates {
(0,0.01084347895832693)(1,0.08692344693076878)(2,0.8440230128801958)(3,0.8912585298835761)
};
\addlegendentry{anomaly-spam}
\addplot[
,color=orange,
mark=square,
]
coordinates {
(0,0.4965471827673785)(1,0.4654541490449361)(2,1.0)(3,1.0)
};
\addlegendentry{dos}
\addplot[
,color=violet,
mark=square,
]
coordinates {
(0,0.11674638054788085)(1,0.5795488684146953)(2,0.9918658399932255)(3,0.99537419916113)
};
\addlegendentry{scan11}
\addplot[
,color=magenta,
mark=square,
]
coordinates {
(0,0.6941061636849043)(1,0.7870802882741675)(2,0.9940958592051669)(3,0.996624607966314)
};
\addlegendentry{scan44}
\addplot[
,color=olive,
mark=square,
]
coordinates {
(0,0.022936081872189377)(1,0.22050524727468962)(2,0.5212617376247476)(3,0.5289858302866763)
};
\addlegendentry{nerisbotnet}
\end{axis}
\end{tikzpicture}}
\resizebox{0.33\linewidth}{!}{\begin{tikzpicture}
\begin{axis}[xlabel={Time performances},
ylabel={Time(s)},
xmin=0,
xmax=3,
ymin=0,
ymax=67.94624433517455,
xtick={0,1,2,3},
xticklabels={Base set, Graph, Graph community, DGC},
ytick={0.9608979225158691,67.94624433517455},
legend pos=south east,
legend style={fill=none},
ymajorgrids=true,
grid style=dashed]
\addplot[
,color=blue,
mark=square,
]
coordinates {
(0,1.0065673828125)(1,0.9608979225158691)(2,1.3837574958801269)(3,1.4973010540008544)
};
\addlegendentry{Prediction time}
\addplot[
,color=red,
mark=square,
]
coordinates {
(0,16.151073837280272)(1,16.861229753494264)(2,50.03711671829224)(3,67.94624433517455)
};
\addlegendentry{Fitting time}
\end{axis}
\end{tikzpicture}}
\end{tabular}
\end{figure}
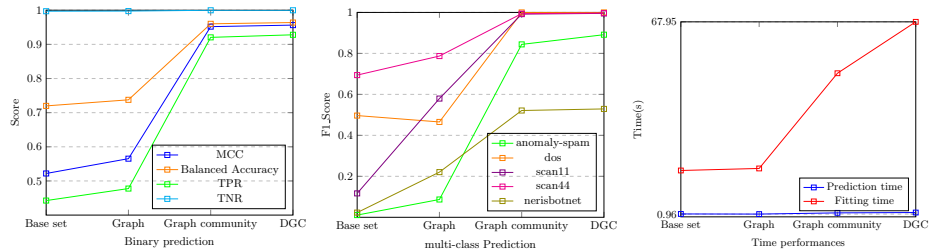


The evaluation highlights the effectiveness of both community and spectral approaches in detecting network threats. The community-based method, referred to as DGC (Dynamic Graph Community) as shown in Fig~\ref{fig:community_evaluation}, demonstrates improved metrics on the UGR16 dataset. For binary predictions, DGC shows Matthews Correlation Coefficient (MCC) of 0.96, Balanced Accuracy of 0.96 and True Positive Rate (TPR) of 0.93, surpassing the baseline with MCC of 0.52 and TPR of 0.44 , as well as graph and graph community models on a 5-folds evaluation. In multi-class predictions, DGC achieves high F1-scores, notably 0.89 for anomaly-spam and 0.99 for both scan44 and scan11 attacks, against respectively 0.01 for anomaly-spam, 0.69 for scan44 and 0.11 for scan11 in the baseline. Time performance analysis reveals prediction times from around 1s to 1.5s, though fitting times increase from 16.15s to 67.95s due to model complexity.

\begin{figure}[H]
\caption{Comparison of spectral graph approaches with baseline on Botnet dataset}
\label{fig:spectral_evaluation}
\begin{tabular}{ccc}
\multicolumn{3}{c}{} \\
\resizebox{0.33\linewidth}{!}{
\begin{tikzpicture}
\begin{axis}[xlabel={Binary prediction},
ylabel={Score},
xmin=0,
xmax=3,
ymin=0,
ymax=1,
xtick={0,1,2,3},
xticklabels={COD, CTS, CTW, SM},
ytick={0, 0.1, 0.2, 0.3, 0.4, 0.5, 0.6, 0.7, 0.8, 0.9, 1},
legend pos=south east,
legend style={fill=none},
ymajorgrids=true,
grid style=dashed]
\addplot[
,color=blue,
mark=square,
]
coordinates {
(0,0.8962486)(1,0.877318)(2,0.980021)(3,0.988675)
};
\addlegendentry{MCC}
\addplot[
,color=orange,
mark=square,
]
coordinates {
(0,0.615637)(1,0.665858)(2,0.988929)(3,0.995543)
};
\addlegendentry{Balanced Accuracy}
\addplot[
,color=green,
mark=square,
]
coordinates {
(0,0.999987)(1,0.999682)(2,0.999740)(3,0.999481)
};
\addlegendentry{TPR}
\addplot[
,color=cyan,
mark=square,
]
coordinates {
(0,0.02414773)(1,0.636364)(2,0.950207)(3,0.983402)
};
\addlegendentry{TNR}
\end{axis}
\end{tikzpicture}}
\resizebox{0.33\linewidth}{!}{
\begin{tikzpicture}
\begin{axis}[xlabel={multi-class Prediction},
ylabel={F1\_Score},
xmin=0,
xmax=3,
ymin=0,
ymax=1,
xtick={0,1,2,3},
xticklabels={COD, CTS, CTW, SM},
ytick={0, 0.1, 0.2, 0.3, 0.4, 0.5, 0.6, 0.7, 0.8, 0.9, 1},
legend pos=south east,
legend style={fill=none},
ymajorgrids=true,
grid style=dashed]
\addplot[
,color=green,
mark=square,
]
coordinates {
(0,1)(1,0.8750)(2,1)(3,1)
};
\addlegendentry{DDoS}
\addplot[
,color=orange,
mark=square,
]
coordinates {
(0,0.8937)(1,0.9966)(2,0.9990)(3,0.9994)
};
\addlegendentry{ScanService}
\addplot[
,color=violet,
mark=square,
]
coordinates {
(0,0.2617)(1,0.8235)(2,0.9953)(3,0.9953)
};
\addlegendentry{OS-Fingerprint}
\addplot[
,color=magenta,
mark=square,
]
coordinates {
(0,0.5333)(1,0.6666)(2,1)(3,1)
};
\addlegendentry{Keylogging}
\end{axis};
\end{tikzpicture}}
\resizebox{0.33\linewidth}{!}{
\begin{tikzpicture}
\begin{axis}[xlabel={Time performances},
ylabel={Time(s)},
xmin=0,
xmax=3,
ymin=0,
ymax=62,
xtick={0,1,2,3},
xticklabels={COD, CTS, CTW, SM},
ytick={0, 62},
legend pos=north east,
legend style={fill=none},
ymajorgrids=true,
grid style=dashed]
\addplot[
,color=blue,
mark=square,
]
coordinates {
(0,0.24)(1,0.013)(2,0.02702)(3,0.0289)
};
\addlegendentry{Prediction time}
\addplot[
,color=red,
mark=square,
]
coordinates {
(0,61.02)(1,1.4543)(2,2.1236)(3,4.5940)
};
\addlegendentry{Fitting time}
\end{axis}
\end{tikzpicture}}
\end{tabular}
\end{figure}
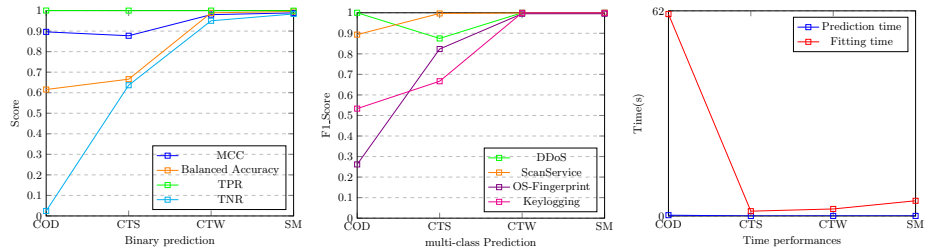
We adopt four distinct approaches to classify network traffic data. The first approach, \textit{Classification on Original Data-logs (COD)}, uses the raw dataset with minimal filtering, excluding only time and sequence-related features to avoid bias from temporal patterns. The second approach, \textit{Classification on Time-Series (CTS)}, introduces a time-series transform mechanism to segment traffic logs data into series of data and extract basic quantitative features like packet counts and rates, aiming to capture evolving traffic behaviors without relying on topological structures. The third approach, \textit{Classification on Time-Windowing (CTW)}, enhances $CTS$ by aggregating data within each time window into new datasets labeled based on the presence of attacks, improving temporal context while still focusing on quantitative attributes. The fourth and final approach, \textit{Spectral Metrics (SM)}, constructs graphs within each time window and splits them into two sub-windows to observe structural changes. It then extracts spectral metrics from Laplacian matrices of these subgraphs, providing topological insights and enabling a graph-based representation of dynamic network behavior for classification. These approaches are detailed in our previous work~\cite{jaber2024graph} for further details.

The spectral approach, represented as SM (Spectral Metrics) as shown in Fig~\ref{fig:spectral_evaluation}, proves highly effective on the Botnet dataset. For binary predictions, SM reaches near-perfect metrics, including an MCC of 0.91 out of approximate of 0.9 for the COD, Balanced Accuracy of 0.97 out of 0.61 for COD, TPR of 0.93, and TNR of 0.99 out 0.01 for COD. In multi-class predictions, SM achieves excellent F1-scores, such as 0.99 for ScanService and 0.91 for DDoS attacks. Despite higher fitting times at 62s for COD, prediction times remain efficient across all configurations.

A comparison evaluation between $SPECTRA$, E-GraphSage and EN-GConv is performed for Botnet \cite{koroniotis2019towards} and TonIoT \cite{moustafa2021new} IoT datasets. The results for binary are shown in tables \ref{fig.binary_toniot}, and \ref{fig.binary_botnet}. The results for multi-classification are shown in Appendix in tables ~\ref{fig.multi-class_toniot}, \ref{fig.multi-class_botnet}. $SPECTRA$ significantly outperforms GCN for BotNet IoT in binary and multi-class detection for all attacks and outperforms them for multi-class detection in 7 cases out of 9 (when considering balanced accuracy) or 8 out of 9 (for F1-score) for TonIoT. For binary classification, it is competitive but second to E-GraphsSage.

\begin{table}[ht]
\caption{Comparison study between E-GraphSage, EN-GConv and SPECTRA over Botnet IoT dataset for binary classification}
\label{fig.binary_toniot}
\begin{tabular}{|l|l|l|l|}
\hline
 & E-GraphSage & EN-GConv & SPECTRA \\ \hline
F1 Score & 0.9888 & 0.3322 & \textbf{0.9944} \\ \hline
Balanced Acc & 0.9796 & 0.8817 & \textbf{0.9971} \\ \hline
MCC & 0.7772 & 0.3892 & \textbf{0.9936} \\ \hline
ADR (\%) & 97.86 & 97.14 & \textbf{99.52} \\ \hline
Precision & \textbf{0.9993} & 0.2004 & 0.9936 \\ \hline
\end{tabular}%
\end{table}

DGC and SM address advanced network security challenges through complementary strengths. DGC excels in analyzing large-scale networks by identifying clusters based on community-driven patterns, which helps detect anomalies like insider threats or advanced persistent threats. On the other hand, SM leverages mathematical rigor to uncover structural insights, identifying hidden substructures such as covert communication channels or botnets.



\section{Impact}

The software opens up avenues for addressing new research questions throughout the pipeline for attack detection. One area of exploration involves graph-based models. Researchers can compare various graph structures and complex network metrics, evaluating their utility in the context of cybersecurity. Additionally, a comparison between graph convolutional models (GCNs), where learning operates directly on graph data, and graph-derived features, where learning operates on tabular data, can reveal insights into their detection capabilities, time performance, complexity, and support for parallelization. 
A promising direction involves exploring how traditional graph metrics and knowledge graphs complement each other to improve detection methods. Traditional metrics—such as degree centrality, betweenness centrality, closeness, and clustering coefficient—help identify structural patterns, including influential nodes or anomalies. Knowledge graphs add contextual semantics, supporting accurate classification of entities and relationships. Combining both approaches strengthens the detection of complex threats and irregular behaviors.

For attack detectors, the software facilitates characterizing detectors for specific types of attacks, defining weak signal analysis to improve detection capabilities, and advancing feature engineering tailored to attack scenarios. Explainability remains a critical focus, providing transparency and interpretability in the detection process.


The software significantly enhances the pursuit of existing research questions in several ways. It enables the automated extraction of derived features, streamlining the process of preparing data for machine learning models. Similarly, it automates the computation of graph metrics, reducing manual effort and increasing efficiency. Moreover, the systematized visualization and evaluation framework provided by the software improves the ability to assess and interpret machine learning approaches for attack detection, enabling more robust and reproducible research outcomes.
\section{Conclusions}
The GPML (Graph Processing for Machine Learning) library provides a robust and scalable solution for addressing the challenges posed by advanced cyber-threats. By leveraging graph-based methodologies, it enables the detection of complex attack patterns, community dynamics, and emerging anomalies in network environments. The library integrates community and spectral graph analysis with powerful Python tools, offering advanced functionalities for temporal graph processing, feature extraction, and visualization. These capabilities not only enhance threat detection but also support deeper exploration of network behaviors, making GPML a valuable resource for researchers and practitioners in cybersecurity. Future work can expand its use for real-time applications and AI-driven threat detection solutions, further advancing network security practices.
\section*{Acknowledgements}
\label{sec:ack}
The authors acknowledge the support of the Région Grand-Est and École Pour l'Informatique et les Techniques Avancées (EPITA) for the joint funding of the XDGMed Project. 

\bibliographystyle{unsrt}
\bibliography{gpml}

\begin{table}[ht]
\caption{Comparison study between E-GraphSage, EN-GConv and SPECTRA over TonIoT dataset for binary classification}
\label{fig.binary_botnet}
\begin{tabular}{|l|l|l|l|}
\hline
 & E-GraphSage & EN-GConv & SPECTRA \\ \hline
F1 Score & 0.966 & \textbf{0.9991} & 0.9977 \\ \hline
Balanced Acc & \textbf{0.9652} & 0.8741 & 0.9493 \\ \hline
MCC & \textbf{0.9623} & 0.1765 & 0.9252 \\ \hline
ADR (\%) & 93.43 & 99.82 & \textbf{99.87} \\ \hline
Precision & \textbf{0.9999} & \textbf{0.9999} & 0.9968 \\ \hline
\end{tabular}%
\end{table}


\begin{table}[H]
\caption{Comparison study between E-GraphSage and SPECTRA over TonIoT dataset for multi-class classification}
\label{fig.multi-class_toniot}
\tiny
\resizebox{\textwidth}{!}{%
\tiny
\scalebox{0.3}{
\begin{tabular}{|l|l|l|l|}
\hline
 &  & E-GraphSage & SPECTRA \\ \hline
DDoS & F1 Score & 0.9823 & \textbf{0.9944} \\ \hline
 & Balanced Acc & 90.839 & \textbf{0.9999} \\ \hline
 & MCC & 0.9760 & \textbf{0.9944} \\ \hline
 & ADR (\%) & 0.9694 & \textbf{100} \\ \hline
 & Precision & \textbf{0.9956} & 0.9889 \\ \hline
DoS & F1 Score & 0.7304 & \textbf{1} \\ \hline
 & Balanced Acc & 0.9220 & \textbf{1} \\ \hline
 & MCC & 0.7006 & \textbf{1} \\ \hline
 & ADR (\%) & 0.9608 & \textbf{100} \\ \hline
 & Precision & 0.5892 & \textbf{1} \\ \hline
Scanning & F1 Score & 0.8549 & \textbf{1} \\ \hline
 & Balanced Acc & 0.8755 & \textbf{1} \\ \hline
 & MCC & 0.8125 & \textbf{1} \\ \hline
 & ADR (\%) & 0.7584 & \textbf{100} \\ \hline
 & Precision & 0.9795 & \textbf{1} \\ \hline
Ransomware & F1 Score & 0.9410 & \textbf{0.9649} \\ \hline
 & Balanced Acc & \textbf{0.9907} & 0.9793 \\ \hline
 & MCC & 90.395 & \textbf{0.9648} \\ \hline
 & ADR (\%) & \textbf{0.9855} & 95.88 \\ \hline
 & Precision & 0.9003 & \textbf{0.9711} \\ \hline
SQL Injection & F1 Score & 0.8282 & \textbf{0.9933} \\ \hline
 & Balanced Acc & 0.9420 & \textbf{0.9962} \\ \hline
 & MCC & 0.8264 & \textbf{0.9933} \\ \hline
 & ADR (\%) & 0.8894 & \textbf{99.24} \\ \hline
 & Precision & 0.7749 & \textbf{0.9943} \\ \hline
Password & F1 Score & 0.9115 & \textbf{0.9869} \\ \hline
 & Balanced Acc & 0.9437 & \textbf{0.9952} \\ \hline
 & MCC & 0.9062 & \textbf{0.9867} \\ \hline
 & ADR (\%) & 0.8915 & \textbf{99.07} \\ \hline
 & Precision & 0.9324 & \textbf{0.9832} \\ \hline
XSS & F1 Score & \textbf{0.9463} & 0.9285 \\ \hline
 & Balanced Acc & \textbf{0.959} & 0.9521 \\ \hline
 & MCC & \textbf{0.9412} & 0.9287 \\ \hline
 & ADR (\%) & \textbf{0.9208} & 90.43 \\ \hline
 & Precision & \textbf{0.9732} & 0.9541 \\ \hline
Backdoor & F1 Score & 0.0787 & \textbf{0.9944} \\ \hline
 & Balanced Acc & 0.5226 & \textbf{0.9972} \\ \hline
 & MCC & 0.0839 & \textbf{0.9942} \\ \hline
 & ADR (\%) & 0.0506 & \textbf{99.46} \\ \hline
 & Precision & 0.1771 & \textbf{0.9942} \\ \hline
MitM & F1 Score & 0.1752 & \textbf{0.9924} \\ \hline
 & Balanced Acc & 0.935 & \textbf{0.9925} \\ \hline
 & MCC & 0.2909 & \textbf{0.9925} \\ \hline
 & ADR (\%) & 0.8743 & \textbf{98.5} \\ \hline
 & Precision & 0.0973 & \textbf{1} \\ \hline
\end{tabular}%
}
}
\end{table}

\begin{table}[]
\caption{Comparison study between E-GraphSage and SPECTRA over Botnet IoT dataset for multi-class classification}
\label{fig.multi-class_botnet}
\tiny
\resizebox{\textwidth}{!}{%
\begin{tabular}{|l|l|l|l|}
\hline
 & \% & E-GraphSage & SPECTRA \\ \hline
DDoS & F1 Score & 0.9999 & \textbf{1} \\ \hline
 & Balanced Acc & 0.9997 & \textbf{1} \\ \hline
 & MCC & 0.9995 & \textbf{1} \\ \hline
 & ADR (\%) & 100 & \textbf{100} \\ \hline
 & Precision & 0.9999 & \textbf{1} \\ \hline
ScanService & F1 Score & 0.9321 & \textbf{0.9986} \\ \hline
 & Balanced Acc & 0.9374 & \textbf{0.9817} \\ \hline
 & MCC & 0.9267 & \textbf{0.9697} \\ \hline
 & ADR (\%) & 87.52 & \textbf{99.89} \\ \hline
 & Precision & 0.9968 & \textbf{0.9982} \\ \hline
OS Fingerprint & F1 Score & 0.7926 & \textbf{0.9953} \\ \hline
 & Balanced Acc & 0.9865 & \textbf{0.9953} \\ \hline
 & MCC & 0.8025 & \textbf{0.9952} \\ \hline
 & ADR (\%) & 98.69 & \textbf{99.07} \\ \hline
 & Precision & 0.6623 & \textbf{1} \\ \hline
Keylogging & F1 Score & 0.8224 & \textbf{1} \\ \hline
 & Balanced Acc & 1 & \textbf{1} \\ \hline
 & MCC & 0.8357 & \textbf{1} \\ \hline
 & ADR (\%) & 100 & \textbf{100} \\ \hline
 & Precision & 0.6984 & \textbf{1} \\ \hline
\end{tabular}%
}
\end{table}

\end{document}